\documentclass[lettersize,journal]{IEEEtran}
\usepackage{amsmath,amsfonts}
\usepackage{algorithmic}
\usepackage{algorithm}
\usepackage{array}
\usepackage[caption=false,font=normalsize,labelfont=sf,textfont=sf]{subfig}
\usepackage{textcomp}
\usepackage{stfloats}
\usepackage{url}
\usepackage{verbatim}
\usepackage{graphicx}
\usepackage{cite}
\usepackage{bm}
\begin{document}

\title{Unsupervised Neural Motion Retargeting \\ for Humanoid Teleoperation}
\author{Satoshi Yagi$^{1}$, Mitsunori Tada$^{2}$, Eiji Uchibe$^{3}$, Suguru Kanoga$^{2}$, Takamitsu Matsubara$^{4}$, and Jun Morimoto$^{1}$
\thanks{$^{1}$Graduate School of Informatics, Kyoto University, Kyoto, Japan}
\thanks{$^{2}$Artificial Intelligence Research Center, National Institute of Advanced Industrial Science and Technology (AIST), Tokyo, Japan}
\thanks{$^{3}$Department of Brain Robot Interface, Computational Neuroscience Labs, Advanced Telecommunications Research Institute International (ATR), Kyoto, Japan}
\thanks{$^{4}$Graduate School of Science Technology, Nara Institute of Science and Technology, Nara, Japan}
}



\maketitle

\begin{abstract}
This study proposes an approach to human-to-humanoid teleoperation using GAN-based online motion retargeting, which obviates the need for the construction of pairwise datasets to identify the relationship between the human and the humanoid kinematics. Consequently, it can be anticipated that our proposed teleoperation system will reduce the complexity and setup requirements typically associated with humanoid controllers, thereby facilitating the development of more accessible and intuitive teleoperation systems for users without robotics knowledge. The experiments demonstrated the efficacy of the proposed method in retargeting a range of upper-body human motions to humanoid, including a body jab motion and a basketball shoot motion. Moreover, the human-in-the-loop teleoperation performance was evaluated by measuring the end-effector position errors between the human and the retargeted humanoid motions. The results demonstrated that the error was comparable to those of conventional motion retargeting methods that require pairwise motion datasets.
Finally, a box pick-and-place task was conducted to demonstrate the usability of the developed humanoid teleoperation system.
\end{abstract}


\section{Introduction}
\label{sec:introduction}

For human-in-the-loop humanoid teleoperation, we need to derive a mapping to convert the captured human motions into humanoid motions. Traditional approaches have used pairwise data sets and inverse kinematics (IK) of the humanoid.
However, finding corresponding humanoid postures for a given large human data set
requires considerable effort.
In addition, the operator must avoid movements that lead to unstable IK computation due to the use of the singular posture or that are outside the humanoid's range of motion.
Furthermore, we need to consider how to exploit the remaining redundancy in the arms and torso \cite{nakanishi2020towards} and define additional constraints \cite{yamane2020kinematic}.

\begin{figure}[t]
\centering
\includegraphics[width=0.9\columnwidth]{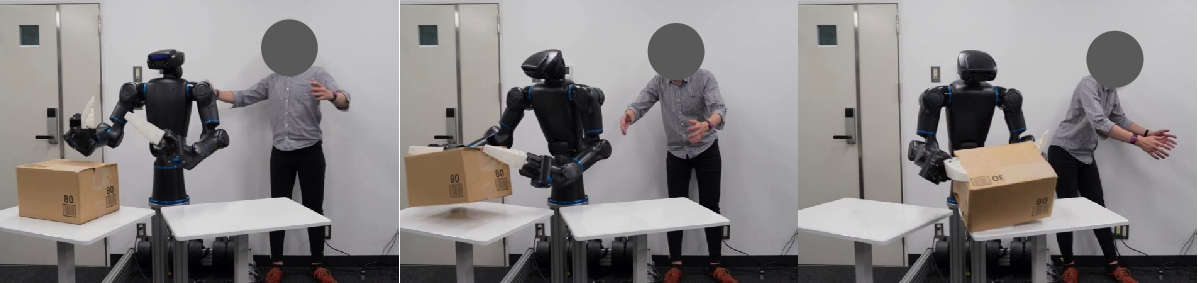}
\caption{Operating a humanoid with the developed controller. Our teleoperation controller does not require paired data sets or common pre-specifications for its learning process to achieve human-to-humanoid motion retargeting.}
\label{fig:figure1}
\end{figure}

Motion retargeting, which is widely used to generate the motion of animated characters, involves transferring a demonstrator's motion to characters with differently structured body shapes \cite{penco2018robust}. 
For example, by attaching markers to a human demonstrator's body and capturing the marker positions, the joint angles of the humanoid are derived to minimize discrepancies in the corresponding body positions.
This manual configuration requires the expertise of specialized designers. In addition, IK-based retargeting often fails when dealing with unexpected motion sequences.

The recent expansion of human motion datasets has spurred the development of data-driven motion retargeting methods. 
In particular, neural network-based approaches are enabling more precise and adaptive motion transfer between different entities. This trend reflects a growing preference for data-driven solutions that can handle a wider range of motions and scenarios without requiring expertise in manual body mapping or user-defined constraints. 
However, extensive data preparation is required, such as labeling and cleaning the training data. This process involves meticulously organizing and refining datasets to ensure accuracy and usability in motion retargeting, which is often a time-consuming and labor-intensive task. In the field of computer animation, several motion retargeting methods using an unpaired dataset have been proposed \cite{villegas2018neural, lim2019pmnet}. Aberman et al. proposed a motion retargeting method using a CycleGAN-based network architecture \cite{aberman2020skeleton}. This method does not require paired data sets. The network encodes the motions of two types of animated characters to obtain a common latent space during training. It then uses this common latent space to facilitate the transformation of motion between characters with different numbers of joints.

To the best of our knowledge, no motion retargeting method has been proposed that uses unsupervised learning without the need for data labeling and without requiring prior specifications for humans and robots to calculate a loss function during training.
Inspired by the idea proposed in \cite{aberman2020skeleton}, this study addresses human-humanoid motion retargeting instead of converting motion sequences between different animated characters using only unpaired datasets. It then allows the human operator to control the humanoid robot in a human-in-the-loop fashion, as shown in Figure \ref{fig:figure1}. 
The network is designed to accommodate variations in body structure, allowing for consistent motion retargeting even when different operators are used. The result is a controller that can be intuitively operated by novice users.

The contributes of this study are as follows:
\begin{itemize}
    \item We proposed a novel skeleton structure for the humanoid model to derive a latent space that efficiently represents the relationships between human and humanoid movements. In addition, we also introduced a novel loss function that explicitly detects the moving joint direction of the humanoid robot. As a result, our proposed method showed better human motion reconstruction performance than the retargeting method \cite{aberman2020skeleton} originally developed for animated characters.

    \item Our method achieves end-effector position errors comparable to conventional IK-based methods without using pairwise data, demonstrating efficient retargeting of human motions.

    \item We developed a teleoperation system and demonstrated its effectiveness by remotely operating a humanoid robot to perform a box pick-and-place task.
\end{itemize}

The remainder of this paper is organized as follows.
Section \ref{sec:relatedresearch} introduces related works.
Section \ref{sec:motionretargeting} explains the proposed model.
Section \ref{sec:humanoidcontroller} describes our teleoperation system for humanoid control.
Section \ref{sec:dataprocessing} outlines the experiment setups focusing on data processing. Section \ref{sec:result} presents the experimental results.
Section \ref{sec:discussion} discusses the characteristics of our proposed model.
Finally, Section \ref{sec:conclusion} concludes this paper.

\section{Related Works}
\label{sec:relatedresearch}

A pioneering study in neural-based human-to-humanoid motion control was conducted by Matsui et al. \cite{matsui2005generating}. The goal of the study was to teleoperate a humanoid equipped with numerous pneumatic actuators in the torso and shoulders. Motion-capture markers were placed in similar locations on both a human and the humanoid, and the actuators drove the internal linkage mechanisms to follow the marker positions as the human moved. Through supervised learning with neural networks, they acquired mappings from human body positions to humanoid joint angles, and from humanoid joint angles to humanoid body positions. By combining these mappings, human-to-humanoid teleoperation was demonstrated.

Stanton et al. reported a similar neural-based motion tracking via supervised learning \cite{stanton2012teleoperation}. 
In their study, the neural network learns to map between the joint rotations of a human movement and the rotations of a robot's motors. To obtain paired motion data before performing teleoperation, a human imitates the motions to be performed by the humanoid.
While this approach eliminates the need for designers to manually map human and humanoid body correspondences, it does require an actual teleoperator to move the body to match the reference motions of the humanoid. The accuracy of motion retargeting is highly dependent on the variety of exemplar motions of the humanoid and the ability of the teleoperator to mimic these motions in time and space. In addition, although this study used a humanoid with a body structure quite similar to humans, some joint movements were fixed to make it easier for the teleoperators to imitate the exemplar motions.

Other highly accurate neural-based motion tracking methods have also been reported \cite{mingon2016human, kim2020c, zhang2022kinematic}. However, these studies require some presettings to be incorporated into the loss function for learning body position correspondences, similar to IK-based methods. For example, the position of the humanoid's elbow must be specified before learning. In addition, although humanoids are used in the research, motion retargeting is limited to arm movements only.

Unsupervised learning methods for motion reconstruction have been proposed to eliminate the burden of data pairing and body part correspondence, which are time-consuming and require specialized knowledge. ImitationNet generates human-like robot motions from human actions using deep metric learning \cite{yan2023imitationnet}. CycleAutoencoder learns motion retargeting by alternately transforming motions through two autoencoders that encode and decode human and robot actions \cite{stanley2021robust}. Both studies achieve motion retargeting without labeling the training data, but the focus of robot control in these research projects is limited to the arms. 
Choi et al. report a motion retargeting method not only for the arms, but also for the upper body of humanoids. This method is in the context of semi-supervised learning and still requires paired data \cite{choi2020nonparametric}.

In the field of animation, Aberman et al. achieved motion retargeting between animated character skeletons without the need for labeling \cite{aberman2020skeleton}. They used a combination of CycleGAN \cite{zhu2017unpaired} and Graph Convolutional Network \cite{kipf2016semi} approaches. Graph neural networks are strong candidates for processing motion in skeletal structures. Unlike traditional convolutional neural networks, graph neural networks process information on a per-node (joint) basis. Their proposed method for skeletal convolution and pooling simplifies different skeletons into a common primal skeleton within a shared latent space. For retargeting purposes, skeletal unpooling reconstructs the source motion onto the target skeleton using deep motion representations of the primal skeleton. However, this method is limited by the fact that the retargeted skeletons typically differ by one link per limb and that joints can move in all directions. When retargeting from humans to robots, the differences in skeletal structures become more pronounced. For example, human arm movements captured by motion capture are represented by a few links, whereas a typical robot arm consists of six or seven links, with each joint moving only in specific directions.
Although Annabi et al. discussed the possibility of applying this method to robotics, their study was limited to validation using only animation datasets, possibly due to the above difficulties \cite{annabi2024unsupervised}.

\section{GAN-based Motion Retargeting}
\label{sec:motionretargeting}
\begin{figure}[htbp]
\centering
\includegraphics[width=0.9\columnwidth]{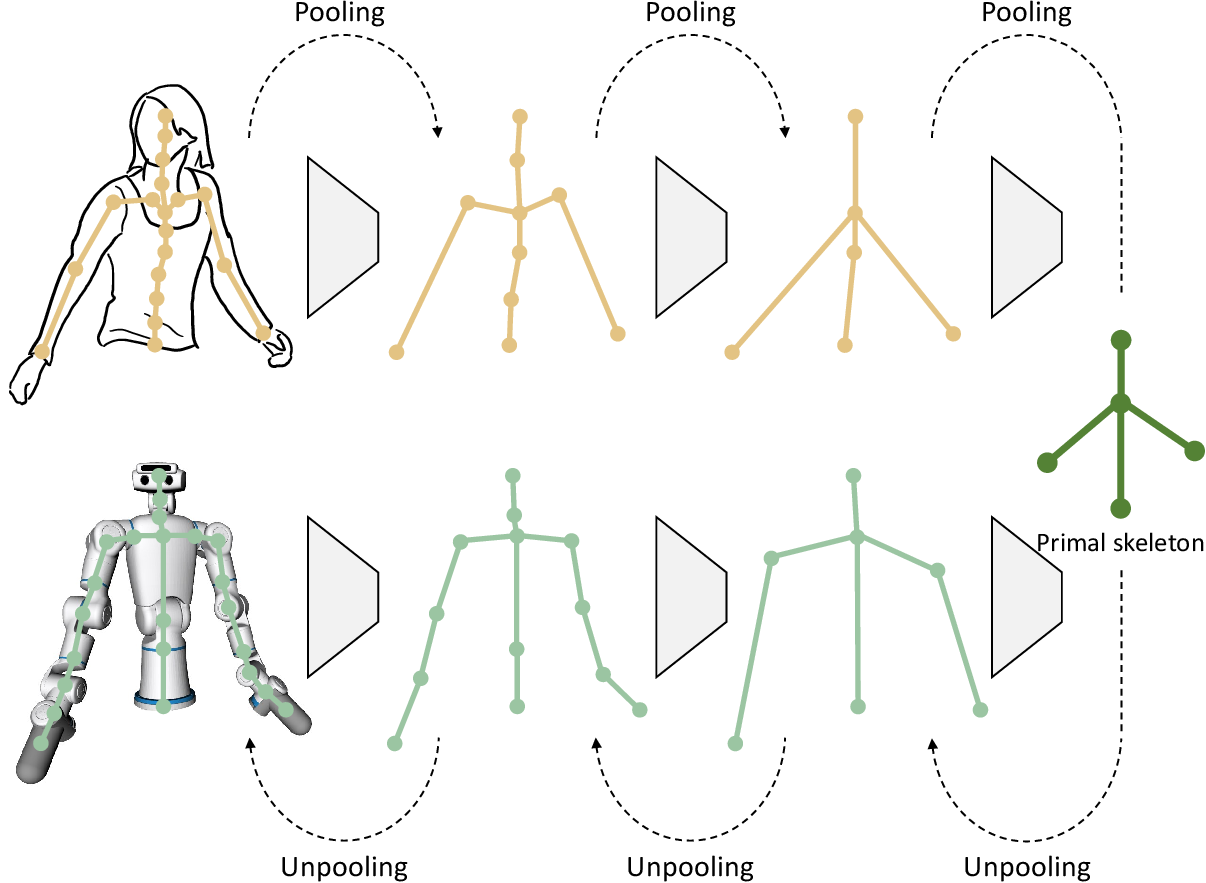}
\caption{Overview of Motion Retargeting: First, the human motion encoders pool two joints of each limb into one. Then, based on the deep motion representations articulated by the original skeleton, the humanoid motion decoders perform an unpooling operation, separating each joint back into two. Ultimately, this process enables the decoders to generate motions that accurately correspond to the actual humanoid skeletal structure.}
\label{fig:figure2}
\end{figure}

In this study, we used separate networks to process static posture and dynamic joint angle data, similar to the approach used in the previous animated character study \cite{aberman2020skeleton}.
This approach allows the sharing of deep motion representations across different skeletal structures. 

Figure \ref{fig:figure2} shows the transformation process of a motion trajectory from a human to a humanoid. As explained above, both a human and a humanoid can be described as topologically equivalent graphs for hierarchical skeletal structures. In other words, both have three limbs, the arms and the head, so they can be represented by a common skeletal structure, like a stick figure. Therefore, it is possible to encode a human motion via this primitive skeletal representation (Figure \ref{fig:figure2}, right side) and decode it as a humanoid motion.

\begin{figure*}[htbp]
\begin{minipage}[b]{0.48\linewidth}
\centering
\includegraphics[width=0.9\columnwidth]{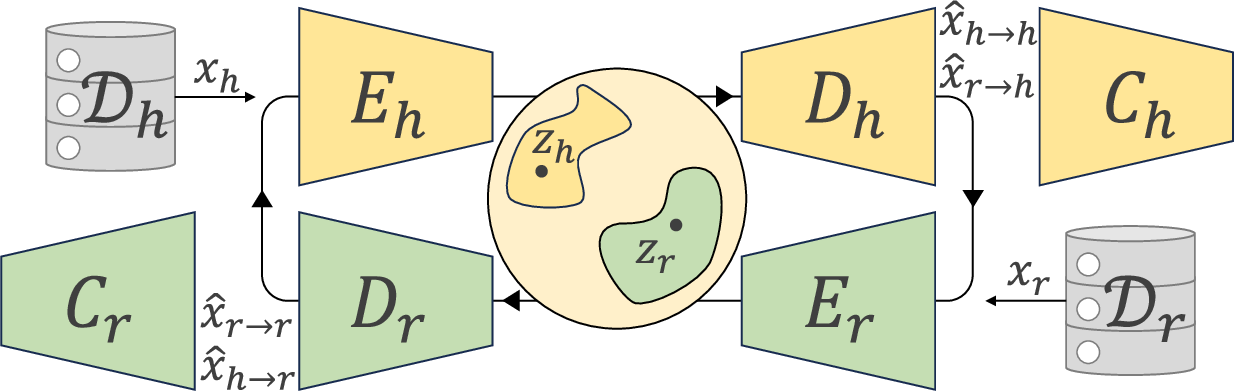}
\\ \centering (a) Training phase.
\end{minipage}
\begin{minipage}[b]{0.48\linewidth}
\centering
\includegraphics[width=0.9\columnwidth]{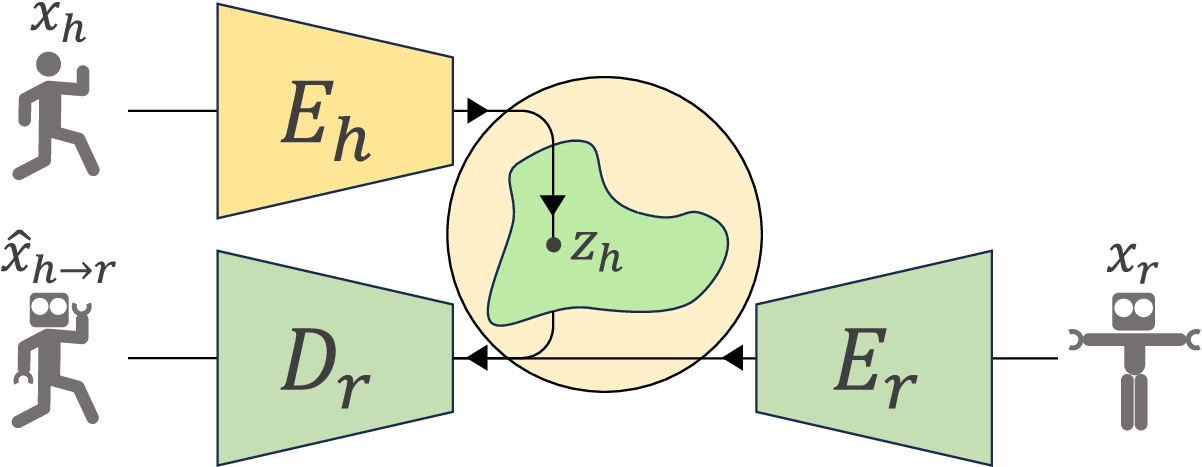}
\\ \centering (b) Teleoperation phase.
\end{minipage}
\caption{Architecture of the network during training and teleoperation: We adopted the same network architecture as in the previous study \cite{aberman2020skeleton}, which was designed for unpaired motion-to-motion translation between two different domains using a CycleGAN approach.
While the previous study achieved motion translation between the same animated characters, this study focuses on motion translation from humans to humanoids, addressing different targets. 
(a) Training phase: During this phase, each generator translates motions from one domain to another, aiming to maintain consistency in motion translations across domains. Simultaneously, each discriminator discriminates between the original motions of its assigned domain and the redirected motions produced by the generator.
(b) Teleoperation phase: In this phase, the network uses the humanoid motion generator developed during training. For motion retargeting, we input source human motion and target human posture information to adapt the humanoid's motions in real time.
}
\label{fig:figure3}
\end{figure*}

Figure \ref{fig:figure3} shows the architecture of the network. 
The architecture includes a CycleGAN designed for unpaired motion-to-motion translation between two different human and humanoid domains.
In the training phase, as shown in figure \ref{fig:figure3}(a),
the network consists of encoders $E_h$ and $E_r$ and decoders $D_h$ and $D_r$ for human and humanoid data, respectively.
Each decoder $D_h$ and $D_r$ is paired with a discriminator $C_h, C_r$, similar to typical generative adversarial networks. 
In our implementation, there are three encoders in series within the encoder $E_{h/r}$, each responsible for consolidating joints from two to one in each limb of the input skeleton, as shown in figure \ref{fig:figure2}. The decoders and discriminators work in a similar way.
The discriminators are trained to distinguish between real training data and data generated by the generator.

We define $\bm{x}_h$ as the original human data from the human dataset $\mathcal{D}_h$, $\bm{z}_h$ as the latent representation output from the encoder $E_h$, $\hat{\bm{x}}_{h \rightarrow h}$ as the reconstructed human data from the decoder $D_h$, and $\hat{\bm{x}}_{h \rightarrow r}$ as the retargeted humanoid data from the decoder $D_r$.
Similarly, $\bm{x}_r$, $\bm{z}_r$, $\hat{\bm{x}}_{r \rightarrow r}$, and $\hat{\bm{x}}_{r \rightarrow h}$ represent the original humanoid data, its latent representation, the reconstructed humanoid data, and the retargeted humanoid data, respectively. Note that the motions of a human/humanoid are determined by its initial posture and the time series of joint angle displacements. Here $\bm{x}$ represents the pair of posture and time series displacement data.

In the teleoperation phase, as shown in figure \ref{fig:figure3} (b), we use the humanoid motion generator developed during training. For motion retargeting, we input the source human motion from $\bm{x}_h$ and the target human posture from $\bm{x}_r$ to adapt the humanoid's motion $\hat{\bm{x}}_{h \rightarrow r}$ in real time.

\begin{figure}[htbp]
\centering
\includegraphics[width=0.9\columnwidth]{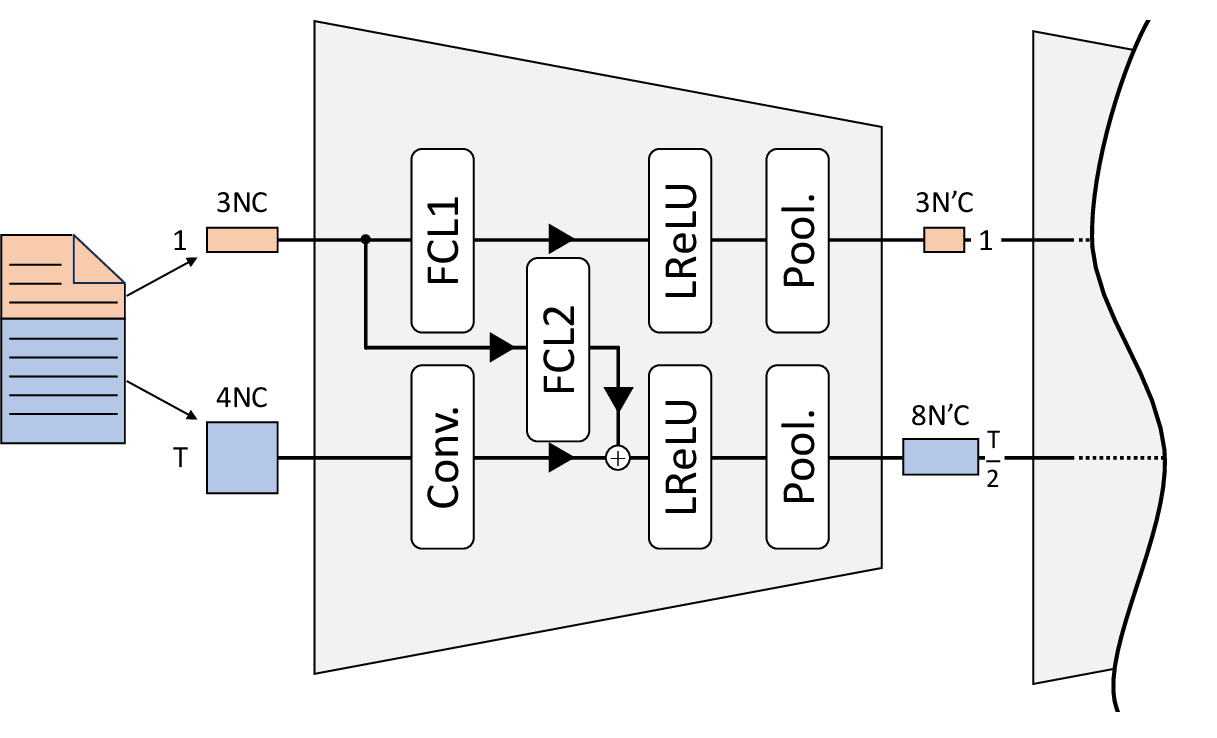}
\caption{Overview of the three-layer encoder architecture: This figure shows one of the three. Inputs (BVH files) are separated into posture (red) and motion (blue) components. Similar to \cite{aberman2020skeleton}, these components are processed in parallel by the upper and lower networks, respectively. The upper network handles posture information and is connected to the lower network, which handles motion, allowing posture data to be integrated during motion encoding.
}
\label{fig:figure4}
\end{figure}

The encoders $E_h$ and $E_r$ each consist of a three-layer structure, figure \ref{fig:figure4} shows one of these layers in detail.
A single movement is determined by the combination of the initial posture (red part) and the time series of joint displacements from the initial posture (blue part). This encoder processes posture and motion data in parallel. The reason for having separate networks for posture and motion is that motions are characterized by skeletal shapes; to extract skeleton-independent motions in latent space, an encoder is needed that compresses motion independently of posture. This allows a common latent motion representation to be obtained even when human operators change. Furthermore, by learning with regularization to make the latent motion representations of humans and humanoids more similar, the network learns the mapping of motions from humans to humanoids and realizes the function of a controller that takes human motions as input and outputs humanoid motions.

In figure \ref{fig:figure4}, the upper network inputs only the initial posture and obtains a deep posture representation compressed by fully connected layers (FCL1). This allows for a deep motion representation independent of the skeleton while preserving skeletal features. The lower network performs graph convolution processing using kernels that support each joint and the time axis (Conv.). The kernel is time-invariant and is not shared between joints. Next, another fully connected layer (FCL2) adds the posture representation to the motion, which is then passed through an activation function (Leaky ReLU).
In this study, the same hyperparameters were used as in the previous study\cite{aberman2020skeleton}.

After the activation function layer, pooling is performed to obtain a deep posture/motion representation. Pooling compresses the results of the convolution layer by thinning them out. Since each limb can move independently, pooling is required for each limb, as in graph convolution networks. In a pooling operation, two vertices from the end to the root vertex are paired and pooled. The pooling operator takes the average. This allows humans or humanoids to obtain the shared latent motion representation while preserving the skeletal structure. For decoders, unpooling performs the opposite operation of pooling, restoring the skeletal structure captured during pooling. In addition, the lower layer of the decoder uses upsampling to double the data along the time axis.

When a human or humanoid skeleton has $N$ joints, excluding the root joint, each joint has a 3D position relative to its parent joint, so the vector representing posture has $3 N$ components. Motion is represented by a quaternion for the $N$ joints over $T$ frames, resulting in a 4$N \ times T$ matrix. With a minibatch size of $B$ and a number of channels of $C$, if we combine the indices 4$N$ and $C$ to 1, ..., 4$NC$, the input size for the upper network is $B \times 3 NC$, and for the convolution layer of the lower network it is $B \times 4NC \times T$. The output size is set to twice the number of channels, resulting in $B \times 8 N'C$ for the deep posture representation and $B \times 8 N'C \times \frac{T}{2}$ for the deep motion representation. Here $N'$ is the number of output skeleton joints.
The deep posture representation is shared with the corresponding decoder.

We used the following loss functions to train the network: $\mathcal{L}_{\text{rec}}$, $\mathcal{L}_{\text{ltc}}$, $\mathcal{L}_{\text{adv}}$, and $\mathcal{L}_{\text{ee}}$. For simplicity, we describe each loss function for the case where human motion data is used as input.

The reconstruction loss $\mathcal{L}_{\text{rec}}^{1 - 3}$ is represented by the equations \ref{eq:rec1}, \ref{eq:rec2}, and \ref{eq:rec3}. Equation \ref{eq:rec1} calculates the mean squared error for all joint time series quaternions $\bm{x}_h$.
Equation \ref{eq:rec2} calculates the mean squared error for all time series joint positions of $\bm{x}_h$, where $f$ is the forward kinematics (FK) function that determines the joint position from the joint rotation (quaternion). We used a coordinate system normalized to a head height of 1.0 to compute the positions.
Unlike a human, a humanoid usually moves its joint in one direction. The equation \ref{eq:rec3} gives a penalty for moving a joint in a direction other than the prescribed one. The function $f_r$ solves the FK with a rotation angle in a predefined direction, zero for other directions. This loss function only applies to the humanoid. In our implementation, when computing the homogeneous transformation matrices for solving IK based on joint rotations, we specifically set the matrix components corresponding to non-rotatable axes to zero. This operation ensures that the transformation matrices accurately reflect the humanoid's mechanical constraints on non-rotating axes. In addition, we confirmed that this implementation stabilizes the training of the network.

\begin{equation}
\mathcal{L}_{\text{rec}}^1 = \mathbb{E}_{\bm{x}_h\sim \mathcal{D}_h}\left[ \left\| \hat{\bm{x}}_{h \rightarrow h} -\bm{x}_h \right\|^2 \right]
\label{eq:rec1}
\end{equation}

\begin{equation}
\mathcal{L}_{\text{rec}}^2 = \mathbb{E}_{\bm{x}_h\sim \mathcal{D}_h}\left[ \left\| f(\hat{\bm{x}}_{h \rightarrow h}) - f(\bm{x}_h) \right\|^2 \right]
\label{eq:rec2}
\end{equation}

\begin{equation}
\mathcal{L}_{\text{rec}}^3 = \mathbb{E}_{\bm{x}_r\sim \mathcal{D}_r}\left[ \left\| f_r(\hat{\bm{x}}_{r \rightarrow r}) - f(\hat{\bm{x}}_{r \rightarrow r}) \right\|^2 \right]
\label{eq:rec3}
\end{equation}

The loss function $\mathcal{L}_{\text{ltc}}$ in equation \ref{eq:ltc} is a cycle consistency loss that helps to acquire the shared latent space between humans and humanoids. Acquiring the shared latent motion representation allows the motion input to the encoder to be transformed into motion by the decoder on the other side. Here $\|\cdot\|_1$ denotes the $L_1$ norm.

\begin{equation}
\mathcal{L}_{\text{ltc}} = \mathbb{E}_{\bm{x}_h\sim \mathcal{D}_h} \left[ \left\| E_r( \hat{\bm{x}}_{h \rightarrow r} ) - E_h( \bm{x}_h )  \right\|_1 \right] 
\label{eq:ltc}
\end{equation}

The loss function $\mathcal{L}_{\text{adv}}$ in equation \ref{eq:adv} is an adversarial loss.
To achieve unsupervised learning with the unpaired human and humanoid motion data, we used the discriminator $C_r$ to estimate whether the real motion $\bm{x}_r$ or the fake motion $\hat{\bm{x}}_{h \rightarrow r}$. $\mathcal{L}_{\text{adv}}$ is the least squares loss function of the least squares generative adversarial network.

{\small
\begin{equation}
\mathcal{L}_{\text{adv}} = 
\mathbb{E}_{\bm{x}_h\sim \mathcal{D}_h} \left[ \left\| C_r(\hat{\bm{x}}_{h \rightarrow r}) \right\|^2 \right]
+\mathbb{E}_{\bm{x}_r\sim \mathcal{D}_r}\left[ \left\| 1 -C_r(\bm{x}_r) \right\|^2 \right] 
\label{eq:adv}
\end{equation}
}

For a humanoid controller, it is critical that the user can intuitively manipulate the retargeting results. In our experience, it is important that the speed scales of the operator's and the humanoid's end effectors match. The loss function $\mathcal{L}_{\text{ee}}$ in equation \ref{eq:ee} imposes a penalty on speed discrepancies that quantifies the positional error of the corresponding end-effectors between humans and humanoids. Here, the function $f_{ee}$ computes the positions of the three end-effectors. In practice, we extracted the end-effector positions from the results of $f$ in the equation \ref{eq:rec2}, where $\Delta$ is the difference from the previous frame and $\Delta t$ is the sampling time.

\begin{equation}
\mathcal{L}_{\text{ee}} = \mathbb{E}_{\bm{x}_h \sim \mathcal{D}_h, \bm{x}_r\sim \mathcal{D}_r}\left[ \left\| \frac{\Delta f_{ee}(\hat{\bm{x}}_{r \rightarrow h})}{\Delta t} - \frac{\Delta f_{ee}(\bm{x}_h)}{\Delta t} \right\|^2 \right] 
\label{eq:ee}
\end{equation}

The complete loss function $\mathcal{L}$ used in training is a combination of the above losses, each multiplied by a coefficient that serves as a hyperparameter. In this study, we used $\mathcal{L} = 5.0\mathcal{L}_{rec}^1 + 2000.0\mathcal{L}_{rec}^2 + 1000.0\mathcal{L}_{rec}^3 +5.0\mathcal{L}_{ltc} +3.5\mathcal{L}_{adv} +225.0\mathcal{L}_{ee}$.

\section{Application to Humanoid Control}
\label{sec:humanoidcontroller}
In this section, we provide an overview of the humanoid controller. This controller only requires the definition of the humanoid’s T-pose for describing its motion in a BVH file. No other pre-settings, such as labeling data or joint pre-specification for the humanoid, are necessary. 
Moreover, since the process of generating motor commands for the humanoid does not involve solving IK, it avoids potential instabilities associated with IK calculations. These instabilities often stem from the operator's inputs, such as singular postures or impractical inputs, which are thus avoided. These features allow even general users, who are not accustomed to motion capture or robot operation, to handle this controller intuitively.

\begin{figure}[htbp]
\centering
\includegraphics[width=0.9\columnwidth]{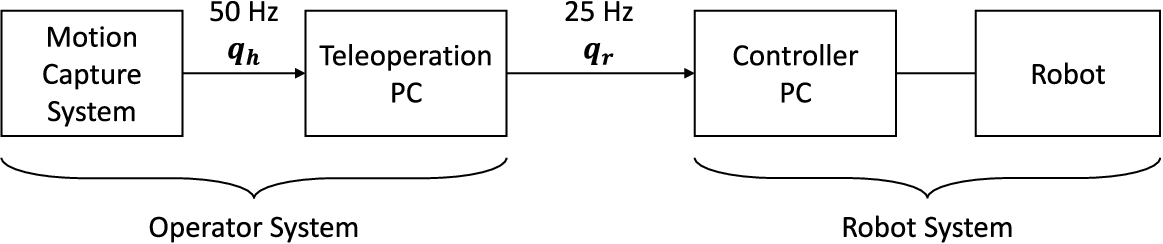}
\caption{System configuration diagram of the humanoid controller. The motion capture system measures human joint angles $q_h$ at 50 Hz. The pre-trained network of the teleoperation PC outputs the desired Torobo joint angles $q_r$ to the controller PC at 25 Hz.}
\label{fig:figure5}
\end{figure}

Figure \ref{fig:figure5} shows the system configuration of the humanoid controller. The operator system consists of a motion capture system and a teleoperation PC (Teleoperation PC). The robot system (Tokyo Robotics Inc.) consists of a PC for controlling Robot (Controller PC) and the humanoid robot Torobo. All modules are connected via wired connections.

For the motion capture system, we used Mocopi (Sony Co.), a wearable wristband motion capture device capable of estimating whole-body motion. Mocopi first estimates the human's posture through a calibration motion (take a step forward) and then captures joint angles $q_h$ at 50 Hz. 

The teleoperation PC is responsible for converting the captured human motion $q_h$ into motion $q_r$ for Torobo. The motion retargeting framework is built using PyTorch and tested on the teleoperation PC that includes an NVIDIA GeForce RTX 4090 GPU (24 GB) alongside an Intel Core i9-13900 CPU. We trained the network's parameters, employing the Adam optimizer with reference to the loss term in the previous section. The training time was approximately 10 hours with the full size dataset described in section \ref{sec:result}, covering 14000 epochs. The pre-trained network inferences every 0.04 seconds using the recent 64 frames of human motion $q_h$ as input $\bm{x}_h$, and outputs only the latest frame of the retargeting motion $\hat{\bm{x}}_{h \rightarrow r}$ as the desired joint angles $q_r$ to the controller PC.

The control PC, equipped with onboard microcomputers and motor drivers, executes the motion control for Torobo based on the received joint angle commands via the ROS (Robot Operating System).

We used the upper-body humanoid robot Torobo. Torobo is a life-sized humanoid with a height of 84 cm. It consists of 22 joints: two joints in the waist (yaw and pitch axes from the base), seven joints in each arm (arranged as pitch, roll, pitch, roll, pitch, yaw, and roll towards the end effector, assuming a T-pose as the initial posture), and two joints in the head (yaw and pitch axes).

\section{Experimental Setups}
\label{sec:dataprocessing}
This chapter outlines the methods used to collect the training data. Collecting motion data for humanoids is more complex. Here, we focus on explaining the humanoid data collection process and the creation of posture information as BVH files for input into the network.

\subsection{Humanoid data}
We used kinesthetic teaching for humanoid motion generation for two main reasons. The first is that it allows anyone, even those without knowledge about robots, to be a teacher because the process is intuitive. The second is that by actually moving the actual humanoid, motions that naturally comply with the hardware specifications can be collected. Thus, by training the network with those motions, it is expected that the trained network will also generate feasible motions that the humanoid can perform.

\begin{figure}[htbp]
\begin{minipage}[b]{0.35\linewidth}
\centering
\includegraphics[width=0.9\columnwidth]{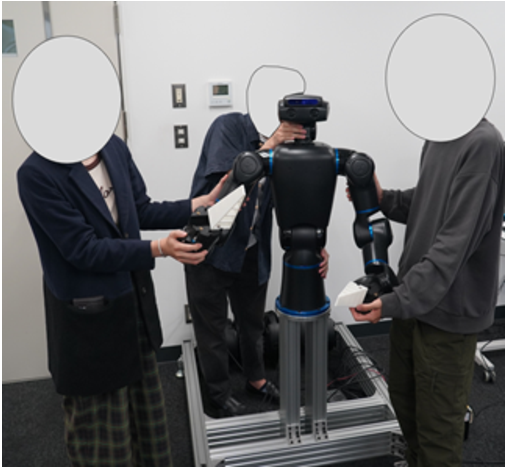}
\\ \centering (a)
\end{minipage}
\begin{minipage}[b]{0.65\linewidth}
\centering
\includegraphics[width=0.9\columnwidth]{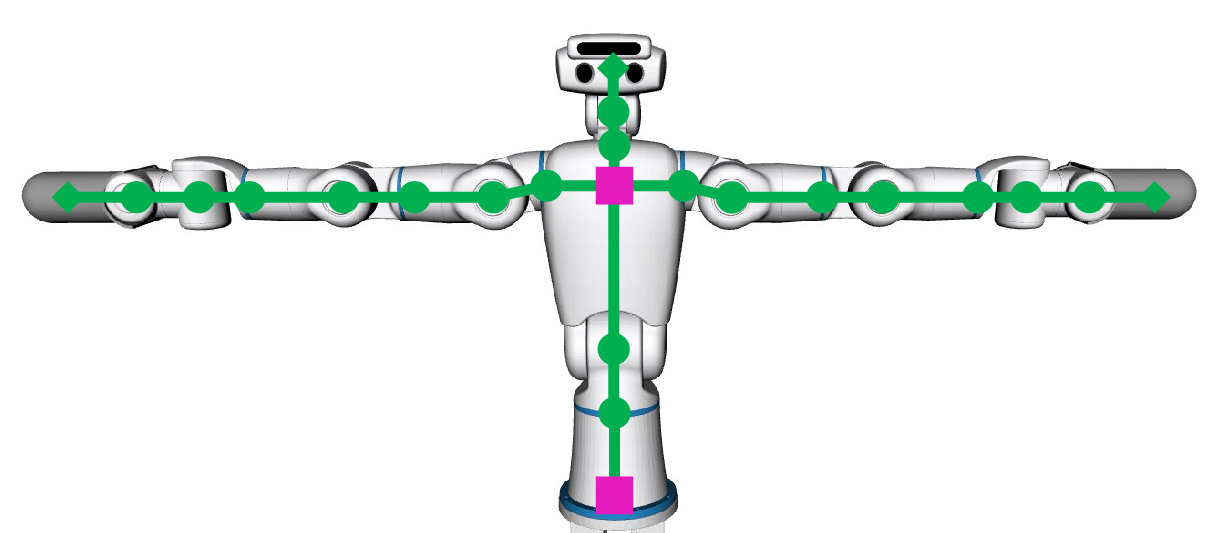}
\\ \centering (b)
\end{minipage}
\caption{Overview of Torobo motion collection. (a) Kinesthetic teaching by three teachers moving Torobo to create training motions. (b) Skeleton structure of Torobo.
}
\label{fig:figure6}
\end{figure}

Figure \ref{fig:figure6} (a) shows the kinesthetic teaching process where three teachers moved Torobo to create motions for training.
Each teacher was assigned the role of teaching the left arm, right arm, or the torso-to-neck region. Teachers pre-planned target motions to teach Torobo. To facilitate easy movement by teachers with minimal force, Torobo was operated in its external force following mode. In this mode, torques counteracting gravity are applied to maintain the current posture. If a teacher moves the body or applies torque to a joint, the joint's torque sensor detects it and drives the joint to follow the applied force. We recorded the sensor readings of the joint angles during Torobo moving at a sampling period of 50 Hz. 

To define the data as BVH files, it is necessary to have posture information.
In the case of humans, commercial motion capture systems automatically generate BVH files from the captured motions and estimated postures through calibration; however, this is not the case for robots.

In creating the posture part of the humanoid's BVH files, the benchmark is to define a skeleton structure in a T-pose that is as close as possible to that of a human. The closer the T-pose between the human and the humanoid, the better the motion retargeting performance.

The concept of this study is to adopt an intermediate skeletal representation, positioned between the actual robot and a human, when defining posture information. This approach does not aim to precisely replicate the robot's skeleton, but rather to facilitate the network's learning of human-robot correspondences.

We defined joints in the BVH file at three specific locations as shown in Figure \ref{fig:figure6} (b): (1) the actual joint positions (green $\circ$), (2) the root and chest joints (purple $\square$), and (3) the end-effectors (green $\diamond$). According to our joint definition protocol, the forward kinematics function $f_r$ in equation \ref{eq:rec3} calculates positions using only the rotational components of the first actual joints (1), while the other joint types (2) and (3) remain fixed and immovable. Here are guidelines for better motion retargeting performance:

(1) When aligning the actual joint structure to match the human T-pose, we ensured even those joints with a bent structure were aligned accordingly.

(2) If a link from the human's chest points upwards or sideways, the corresponding humanoid link should mirror this orientation, either pointing upwards or sideways, respectively.

(3) For end effectors, positioning the joint at a distance equal to the length of its preceding parent link had been shown to improve performance.

\subsection{Human data}
Human data was acquired using the previously mentioned portable motion capture device, Mocopi. While the network can accept data from people with different postures, the number of joints must be the same. Therefore, it is desirable to collect training data using the same motion capture device during operation. No specific rules were set for motion collection, and a variety of motions were prepared, including office life, sports, and dance. 

\subsection{Additional datasets}
Additionally, we used retargeted Torobo motions and source human motions reported by Kang et al. \cite{kang2023curriculum} for evaluation. This method retargeted human motions from the KIT bimanual manipulation dataset \cite{KrebsMeixner2021} to Torobo using an IK-based method \cite{ayusawa2017motion}.

The dataset primarily consisted of kitchen manipulation tasks, covering eleven motion categories: close, cut, mix, open, peel, pour, roll out, scoop, sweep, transfer, and wipe, totaling 93 motions. We added 82 motions (excluding ``open" and ``wipe") for training, and the remaining 11 motions for evaluation.

In summary, we prepared 389,157 frames of Torobo motions (approximately 2.2 hours) and 281,155 frames of human motions (approximately 1.6 hours) to train the network.

\section{Results}
\label{sec:result}
In this section, we show the effectiveness of our proposed humanoid controller from the following three perspectives: 

\begin{itemize}
\item Motion retargeting performance learned by the network.
\item Comparison of our method with the IK-based approach.
\item Implementation of actual teleoperation.
\end{itemize}

\subsection{Motion retargeting performance learned by the network}
Firstly we assessed the performance of our neural-based motion retargeting. Since there is no single correct corespondance data, like ground truth, for robot motion retargeted from human motion, we used a quantitative metric that a retargeted robot motion back to a human and measured the deviation from the original one.

For the evaluation, we first retargeted human motions to Torobo, and then further retargeted these motions back to human motions again. We refer to those reconfigured motions $\hat{\bm{x}}_{h \rightarrow r \rightarrow h}$ as ``cycle-reconstructed motions", to distinguish from retargeted motions $\hat{\bm{x}}_{h \rightarrow r}$.

We prepared additional ten types of human motions $\hat{\bm{x}}_{h \rightarrow r}$, totaling 6380 frames, which were not included in the training data. After the network retargeted the motions to Torobo $\hat{\bm{x}}_{h \rightarrow r}$, we again retargeted them back to human motions $\hat{\bm{x}}_{h \rightarrow r \rightarrow h}$. 

\begin{figure*}[htbp]
\centering
\includegraphics[width=0.9\textwidth]{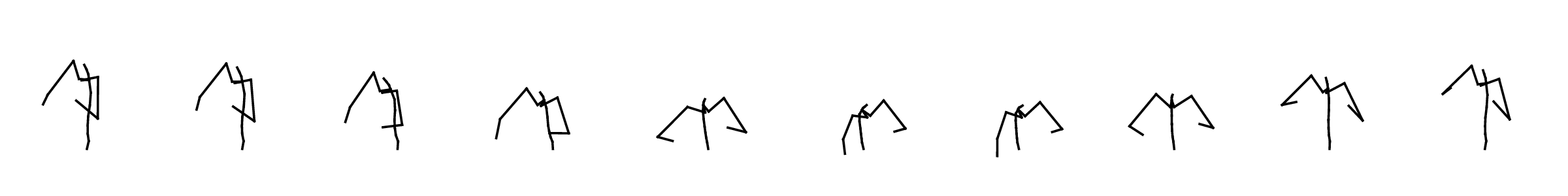}
\includegraphics[width=0.9\textwidth]{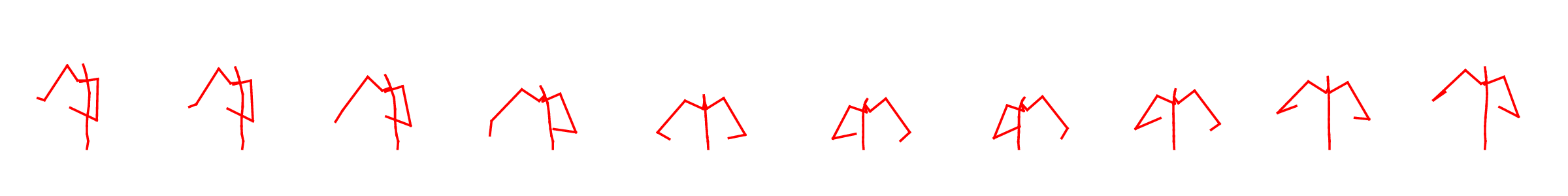}
\includegraphics[width=0.9\textwidth]{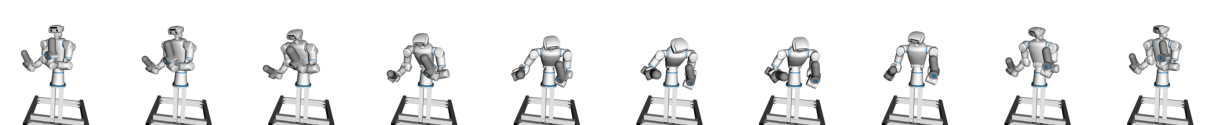}
\bigskip \\ \centering(a) Body jab cross\\
\includegraphics[width=0.9\textwidth]{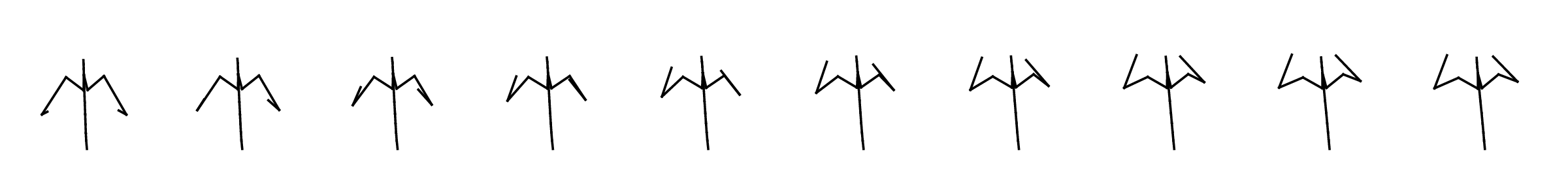}
\includegraphics[width=0.9\textwidth]{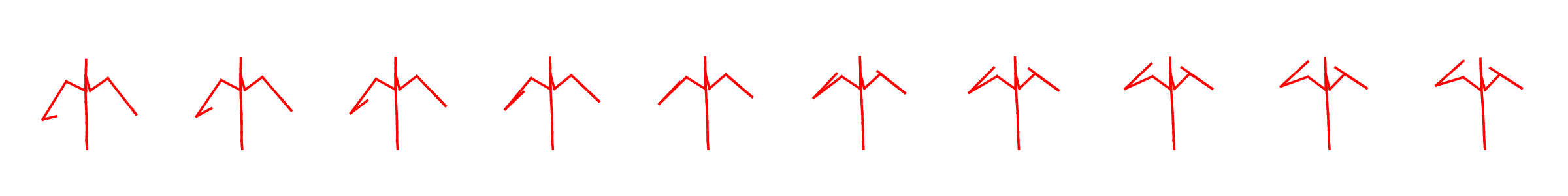}
\includegraphics[width=0.9\textwidth]{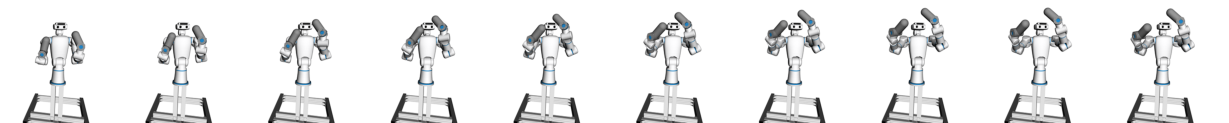}
\bigskip \\ \centering(d) Basketball shoot \\
\caption{Output of the retargeting network: (a) Body jab cross motion with a frame interval of 160 ms, and (b) Basketball shoot motion with a frame interval of 120 ms. The motions are arranged from left to right in time-series order. The first row shows the original human motion in black. The second row shows the cycle-reconstructed motion in red. The third row shows the retargeted Torobo's motion in a simulator.}
\label{fig:figure10}
\end{figure*}

Figure \ref{fig:figure10} shows the representative results; (a) a body jab cross motion, frame interval was 160 ms and (b) a basketball shooting motion, frame interval was 120 ms. The motions are arranged from left to right in time-series order. 
The source human motions are shown in the first row in black.
The cycle-reconstructed motions are shown in the first row in red.
We can see that the network successfully learned to retarget motions from humans to Torobo.

The average position error of all joints between ten types of original motions and the cycle-reconstructed motions was 14.7 mm. Note that we set the coordinate system origin at the location of the root joint (the base of the robot), so this average calculation does not include the root joint. This exclusion also applies to subsequent average position error calculations.

\begin{figure}[htbp]
\centering
\includegraphics[width=0.48\textwidth]{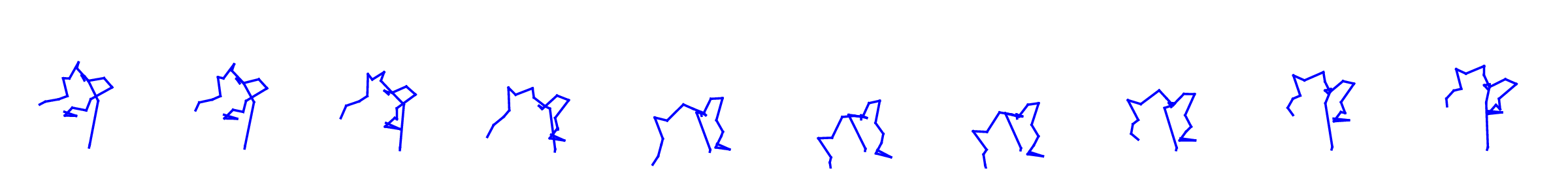}
\centering(a) Body jab cross\\
\includegraphics[width=0.48\textwidth]{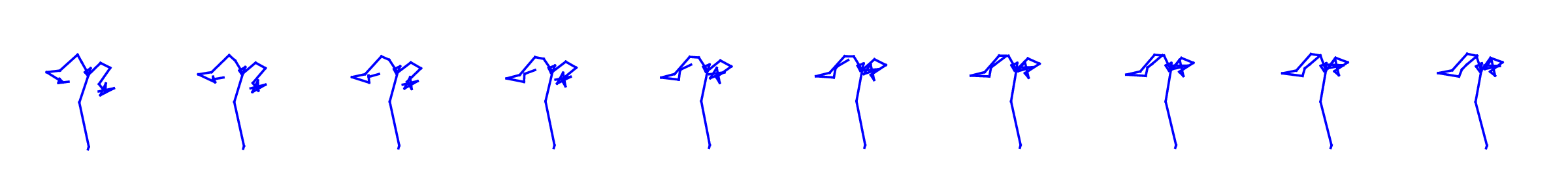}
\centering(d) Basketball shoot \\
\caption{Motion retargeting results using the previous study's method \cite{aberman2020skeleton}.}
\label{fig:figure10-1}
\end{figure}

Figure \ref{fig:figure10-1} shows Torobo's retargeted motions based on the previous study \cite{aberman2020skeleton}. The network was trained for the same number of epochs (14000). The motions are depicted at the same intervals as in Figure \ref{fig:figure10}. Each link does not reflect the actual joint structure, allowing movement in all directions. Compared to the skeleton used in the previous study, Torobo has more arm joints, resulting in unsuccessful retargeting. The average position error of all joints between the original motions and the cycle-reconstructed motions across ten types was 32.3 mm. Therefore, our proposed method demonstrated twice the performance of the previous method, which was originally developed for animated characters.

\subsection{Comparison of our method with the IK-based approach}
Secondly, we demonstrate how closely our retargeted robot motions resemble the previous IK-based retargeted motions \cite{kang2023curriculum} described in Chapter \ref{sec:dataprocessing}.

\begin{figure*}[htbp]
\centering
\includegraphics[width=0.9\textwidth]{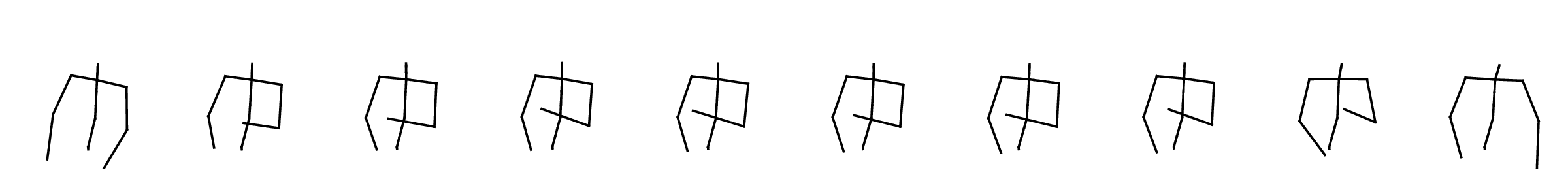}
\includegraphics[width=0.9\textwidth]{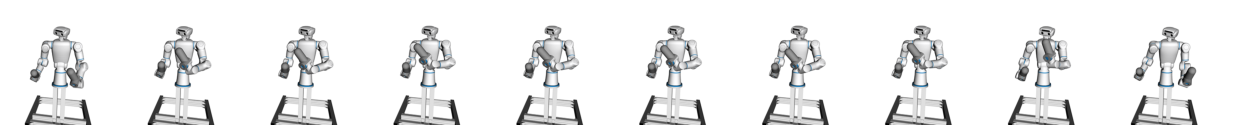}
\includegraphics[width=0.9\textwidth]{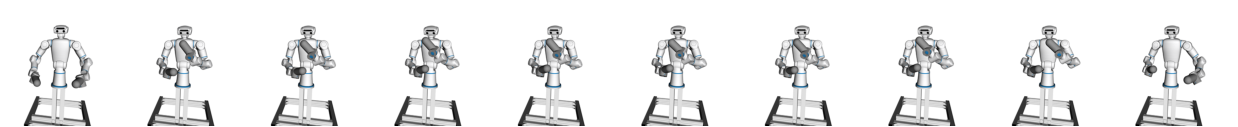}
\bigskip \\ \centering(a) Open\\
\includegraphics[width=0.9\textwidth]{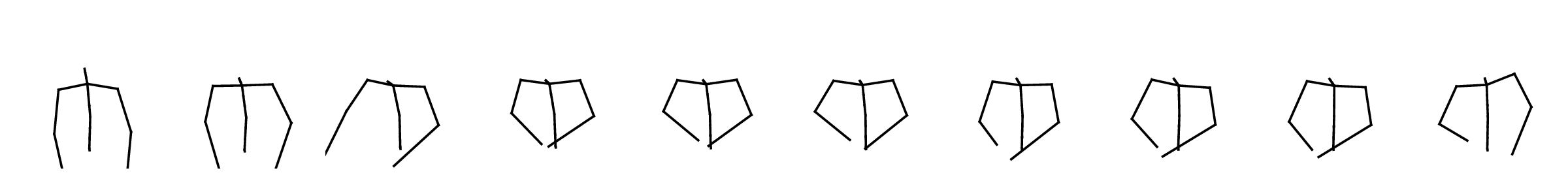}
\includegraphics[width=0.9\textwidth]{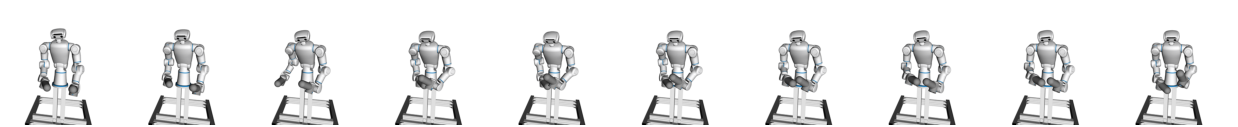}
\includegraphics[width=0.9\textwidth]{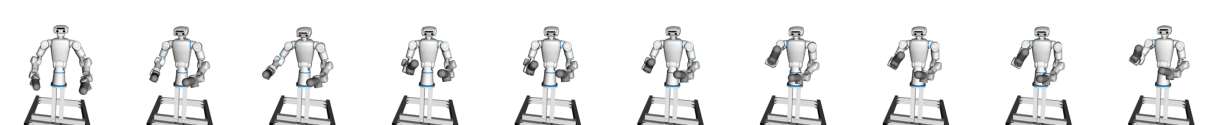}
\bigskip  \\ \centering(b) Wipe\\
\caption{Comparison of retargeted motions of KIT motion datasets: (a) Open motion with a frame interval of 1.4 s, and (b) Wipe motion with a frame interval of 1.2 s. The motions are arranged from left to right in time-series order. The first row shows the original human motion in black. The second and third rows show ours and IK-based in a simulator, respectively.}
\label{fig:figure11}
\end{figure*}

Figure \ref{fig:figure11} shows the comparison of our retargeted motions and the IK-based retargeted motions with the motions from the KIT motion datasets: (a) Open, with a frame interval of 1.4 s, and (b) Wipe, with a frame interval of 1.2 s. The motions are arranged from left to right in time-series order. The first row shows the original human motion in black. The second and third rows show our method and the IK-based method in a simulator, respectively. The IK-based method begins from the initial posture in the first frame. Note that the motions from the KIT motion datasets are made at a slower speed to be easily mimicked by robots, so we showed the results with a larger frame interval for better readability compared to Figure \ref{fig:figure10}.

The average position errors of all joints between our method and the IK-based method were 9.4 cm for the open motion and 15.4 cm for the wipe motion. Table \ref{tab:table1} shows the average position errors of end-effectors in the open/wipe motions compared to the KIT human motions. Our results indicate that our network has similar joint position errors to the IK-based method, despite not requiring labeling for the human and humanoid data.
While our network does not aim to make the position error of each end-effector exactly zero, it achieves positions that are approximately consistent with those of humans.


\begin{table}[htbp]
\caption{The average position errors [cm] of the three end-effectors between the KIT human motions and the motions obtained through our method (left) and the IK-based method (right).}
\centering
\begin{tabular}{|c|ccc|ccc|}\hline
\multicolumn{1}{|c|}{} & \multicolumn{3}{c|}{Ours} & \multicolumn{3}{c|}{IK-based}\\
\cline{2-7}
 & Head & L. Hand & R. Hand & Head & L. Hand & R. Hand \\ \hline
 Open & 10.7 & 26.1 & 17.8 & 11.2 & 24.2 & 22.5 \\
 Wipe & 3.8 & 31.9 & 16.2 & 12.9 & 32.2 & 36.5 \\ \hline
\end{tabular}
\label{tab:table1}
\end{table}

\subsection{Implementation of actual teleoperation.}
\begin{figure*}[htbp]
\centering
\includegraphics[width=0.9\textwidth]{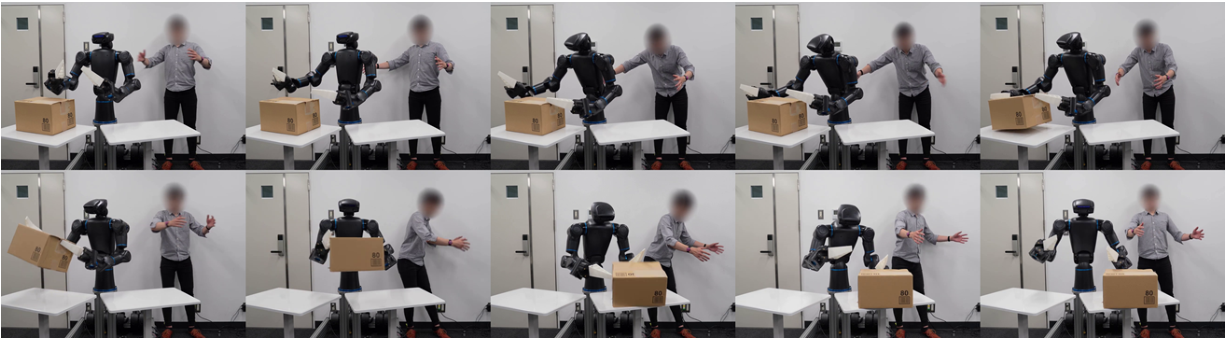}
\caption{Actual teleoperation of the pick-and-place task, arranged from the top left to the bottom right at 1.3-second intervals.}
\label{fig:figure12}
\end{figure*}

We performed a pick-and-place task using teleoperation with Torobo, as shown in Figure \ref{fig:figure12}.
Figure \ref{fig:figure12} displays the sequence of motions at 1.3-second intervals, arranged from top left to bottom right. The operator stood behind Torobo and visually guided its motions.

In this task, Torobo picked up a box from the desk in front of its right side and placed it on another desk in front of its left side. Since the box was placed outside the workspace of Torobo's arms, it had to bend its waist and stretch out its arms in a human-like manner to pick up the box.

During frames 1-5, Torobo twisted and bent forward to pick up the box. During frames 6-8, Torobo placed the box on the adjacent desk. After frame 9, Torobo returned to its initial posture.

\section{Discussion}
\label{sec:discussion}
In this study, we introduced a novel method for humanoid teleoperation, specifically focusing on a motion retargeting approach that does not require labeled training data or pre-specification of joint structures for humanoids. Through experiments, we confirmed our motion retargeting performance using collected human motions and motions from the KIT motion dataset, comparing it to the IK-based method. We also validated our teleoperation system by performing a pick-and-place task.

Preparing the necessary dataset for training remains a challenge. What happens if an insufficient amount of data is available? To explore this, we created medium and small-sized datasets by randomly selecting human and humanoid motions from the full-size dataset used in our experiments. We then trained the network again using nine combinations of three different dataset sizes for both humans and humanoids.

Table \ref{tab:table2} shows the average joint position errors for original and cycle-reconstructed motions learned by the network with each of the three dataset sizes. Interestingly, the errors remained almost the same for all dataset sizes except for the full-size datasets of both human and Torobo data. This indicates that the performance of motion retargeting does not proportionally increase with the amount of data. Instead, a certain threshold of data is necessary to achieve effective teleoperation. 

Therefore, future research directions include developing methods for estimating the required amount of data for training and improving learning efficiency through data augmentation.

\begin{table}[htbp]
\caption{Comparison of average joint position errors [mm] with full, medium, and small datasets. Human datasets are vertical; Torobo datasets are horizontal. Frame counts (in thousands) are in parentheses.}
\centering
\begin{tabular}{|c|ccc|}\hline
& Full (389k) & Middle (200k) & Small (100k) \\ \hline
 Full (281k) &14.7 &31.8 &31.5 \\
 Middle (130k) & 31.2 & 34.3 & 33.5\\
 Small (65k) & 29.0 & 34.9 & 34.4\\ \hline
\end{tabular}
\label{tab:table2}
\end{table}

One of the limitations of this study is that it does not consider the orientation of the end-effectors. For the robot to perform dexterous manipulation like a human, the orientation of the end-effectors is crucial. Furthermore, depending on the task, it may be necessary to provide force feedback to the operator or to have the robot understand the operator's intentions and execute tasks semi-autonomously, requiring additional systems.

\section{Conclusion}
\label{sec:conclusion}
This paper presents a novel approach to online human-to-humanoid motion retargeting for teleoperation that eliminates the need for unpaired datasets and joint pre-specification. By enabling humans to serve as intuitive teleoperation controllers, our method reduces the complexity and setup requirements typically associated with humanoid controllers, paving the way for more accessible and user-friendly teleoperation systems. 

The experiments demonstrated the network's motion retargeting performance and the results of actual teleoperation tasks. Future work will focus on refining the network's performance and exploring additional applications to further enhance the usability and versatility of humanoid robots in various environments.

\section*{Acknowledgment}
This work was supported by JST-Mirai Program, Grant Number: JPMJMI21B1, JST Moonshot R\&D, Grant Number: JPMJMS223B-3,  and JSPS KAKENHI Grant Number: JP22H03669 and JP22K21275.


\bibliographystyle{IEEEtran}

\end{document}